\documentclass[10pt, a4paper]{article}

\usepackage[final]{lrec2026} 

\usepackage{booktabs}
\usepackage{amsmath}
\usepackage{xcolor}
\usepackage{tcolorbox}
\usepackage{adjustbox}
\usepackage{fancyvrb}
\usepackage{fvextra}
\usepackage{enumitem}
\usepackage{array}
\usepackage{url}

\usepackage{breakurl}

\title{Synthetic Function Demonstrations Improve Generation \\ in Low-Resource Programming Languages}

\name{%
  \parbox[c]{0.9\linewidth}{\centering
    {Nick M\raisebox{0.2ex}{c}Kenna$^{\dag}$\textsuperscript{1}, Xinnuo Xu$^{\dag}$, Jack Williams$^{\dag}$, Nick Wilson$^{\dag}$\\
    Benjamin Van Durme$^{\ddag}$, Christian Poelitz$^{\dag}$}
  }%
} 

\address{$^{\dag}$Microsoft Research, $^{\ddag}$Microsoft \\
         nmckenna@microsoft.com, cpoelitz@microsoft.com\\}

\abstract{
A key consideration when training an LLM is whether the target language is more or less resourced, for example English compared to Welsh, or Python compared to Excel. Typical training data for programming languages consists of real program demonstrations coupled with explanatory human-written comments. In this work we present a novel approach to the creation of such data for low resource programming languages, which lack naturally occurring data. Our process generates synthetic, textbook-quality demonstrations of how to use library functions, which we show makes for good model finetuning data. We demonstrate in an example domain of Excel Formulas. First, we collate language documentation, then we use this to augment a powerful teacher model which generates synthetic training data, and finally finetune student models on the demonstrations. Our technique improves student performance on 2 question-answering datasets: WikiTQ and TAT-QA. We also show advantages of finetuning over standard RAG approaches, which can offer only modest improvement due to the unfamiliarity of the target domain to student models.
 \\ \newline  }

\begin{document}
\maketitleabstract
\footnotetext[1]{Now at GitHub Applied Science.}
\setcounter{footnote}{1}

\section{Introduction}

The process of pretraining LLMs on code has improved significantly, supporting high accuracy single-shot generation for general-purpose programming languages (PLs) like Python, when prompted with a variety of natural- and programming language inputs \cite{openai2024gpt4technicalreport, abdin2024phi3technicalreporthighly, qwen2025qwen25technicalreport}. These high-resource PLs (HRPLs) profit due to the abundance of training data, which is often human-annotated with explanatory comments \cite{Kocetkov2022TheStack, lozhkov2024starcoder2stackv2}. Comparatively, LLM generations in Low-Resource Programming Languages (LRPLs) are of significantly worse quality due to the limited data available to train models \cite{joel2024survey}. In this paper, we show that \textit{synthetically generated} data can be used to improve performance when generating in an LRPL.
\footnote{Code, prompts, and data are available at: \\ \hyperlink{https://github.com/microsoft/Synthetic-Function-Demonstrations}{https://github.com/microsoft/Synthetic-Function-Demonstrations}.}

Previous studies directly augment real-world data in high-resource languages using SoTA LLMs \cite{honovich-etal-2023-unnatural, yoo-etal-2021-gpt3mix-leveraging}. However, by definition even SoTA LLMs such as GPT-4o struggle to generate code in low-resource languages \cite{joel2024survey}, making it challenging to directly generate new training data. This makes such previous methods, and more general knowledge distillation approaches, less appropriate for LRPLs.

\begin{figure}[h]
    \centering
    \includegraphics[width=1.0\linewidth, trim=0 8cm 0 0, clip]{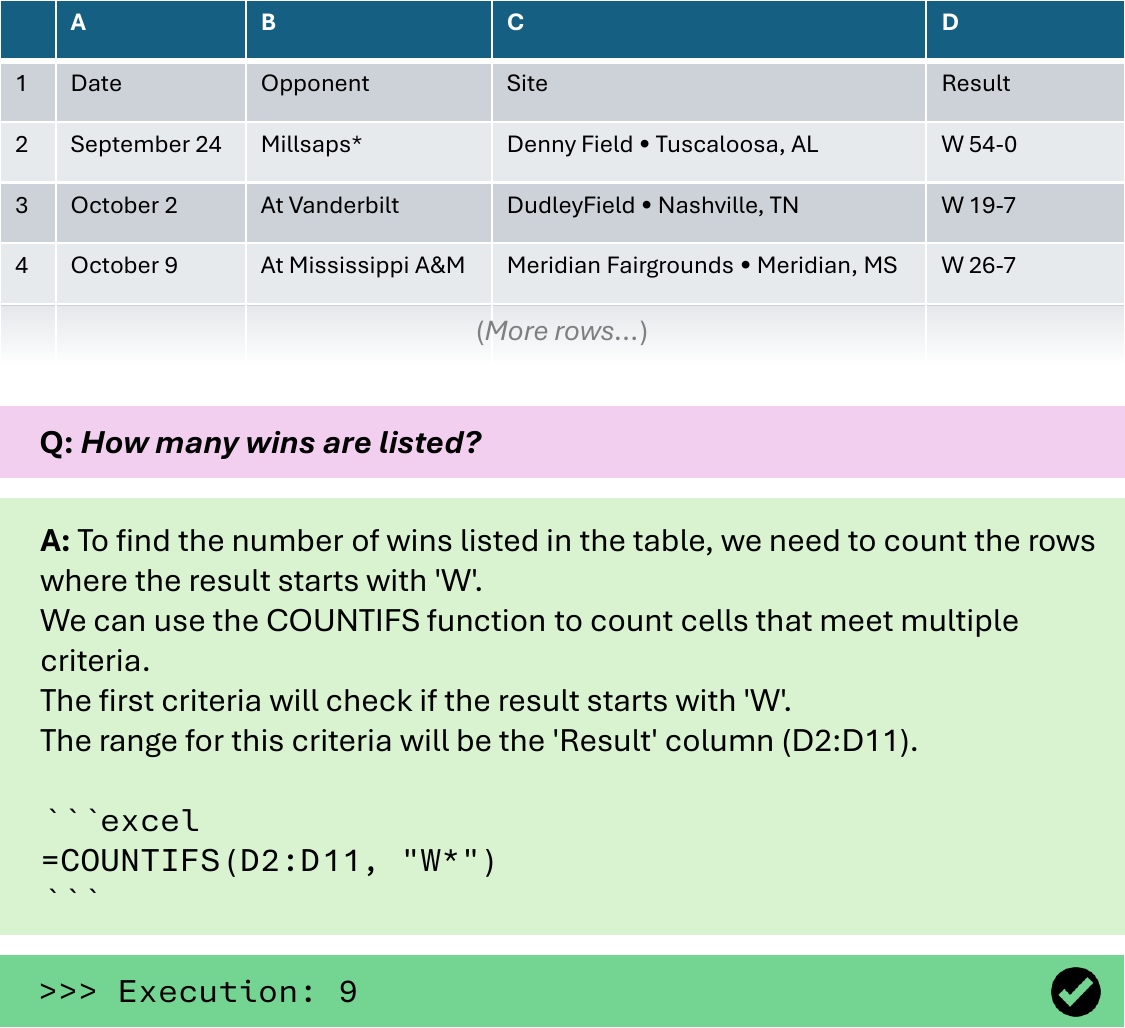}
    \caption{Finetuning on synthetic data improves adaptation to the low-resource domain. Qwen2.5-coder 3B learns to read the table and compose a formula with flexible string formatting, executing to the answer.}
    \label{fig:hero}
\end{figure}

Fortunately, many LRPLs have public, human-written \textit{documentation} detailing language specifications and core library function usage. In this work, we instead generate textbook-quality demonstrations 
based on this documentation \cite{gunasekar2023textbooksneed}.
Generating demonstrations that span the vast space of possible programs is also a challenging task. However, modern LLMs have already learned basic principals of programming and problem-solving by training on plentiful data from HRPLs, and we hypothesize that these concepts will transfer when adapting to a new programming language domain. 
We instead restrict the scope of generated data to the target LRPL's standard library, generating rich demonstrations of each function.

Our goal is to develop data for enhancing model generations which (1) efficiently captures the syntax and semantics of the target LRPL using a minimal amount of data, and (2) leverages models' existing knowledge of general programming concepts, learned from pretraining on large HRPL data, to reason through complex problems in the target domain using the new functions.

Data sparsity is a difficult problem, especially for LRPLs like niche Domain-Specific Languages (DSLs). 
In this paper we focus on Excel Formulas \cite{zhao-etal-2024-nl2formula}: though widely used by millions, formulas are considered a low resource language because of the lack of large-scale natural data sufficient for LLM training \citep{Kocetkov2022TheStack}.
We examine the task of question-answering on tabular data through augmentation strategies on student models to improve generation in Excel formulas. 

This work primarily proposes a general protocol for efficiently generating synthetic training data for a target LRPL. The data robustly demonstrates the usage of library functions in natural contexts and with assistive explanations, tuned to improve model performance on a task. We 
demonstrate efficacy of the approach by augmenting teacher model GPT-4o (which we show is not itself Excel-specialized), and two student LLM model families, Qwen 2.5 and Llama 2 (chosen for their open weights and diverse family of pretrained models) on two table-based QA datasets, WikiTQ and TAT-QA. Within each model family, we also test the effects of model scale, and also dedicated code-pretraining vs. not.


Our contributions are as follows:
\begin{enumerate}



    \item We define a protocol for synthesizing domain-targeted training data to teach an LRPL to student models. Generation is done by augmenting a teacher LLM \textit{without} strong understanding of the target domain simply by retrieving rich function documentation and sampling a general data context for realistic grounding.
    
    \item We show that finetuning on our synthetic examples improves table-QA accuracy by more than 10\% for most student models. Also, we show that code pre-training in HRPLs does not directly improve Excel generation, but rather makes domain adaptation through finetuning on Excel more effective.
    
    
    \item We show that finetuning is necessary in such an unfamiliar LRPL as Excel by comparing with typical RAG-only techniques, which do not significantly improve generation accuracy, even with access to oracle-retrieved information.
\end{enumerate}

    
    

\section{Background}

Code has become integral to general LLM pretraining data and techniques \citep{deepseekai2025deepseekr1}, and to logical reasoning at test-time through code generation \citep{madaan-etal-2022-language, puerto-etal-2024-code} and tool invocation \citep{schick2023toolformer}.

Like natural languages, programming languages suffer from a large disparity in resources \cite{magueresse2020lowresource, lozhkov2024starcoder2stackv2}: generation in low-resource languages is demonstrably worse than in high-resource ones, even by the same model \cite{joel2024survey}. In particular, Excel formula language is a unique LRPL because it is used by so many people, yet there exists little natural training data for formulas. This results in poor generation performance even from top models, which we also document in this work (\S\ref{sec:datasets}). 

While \citet{payan-etal-2023-instructexcel} produce an Excel-specific benchmark which could be used for training or testing, it contains mostly OfficeScript, a different Excel language distinct from Formulas. In contrast, this work presents a method for generating arbitrary amounts of synthetic data tuned to a target domain, and we demonstrate this for Excel formulas. Another study, \citet{zhao-etal-2024-nl2formula}, converts SQL queries (which are abundant) into Excel ones, but this data is unnatural and biased to SQL capabilities and use-cases. In this work we generate synthetic data arising from in-domain documentation and spanning the native capabilities that are described there.

\section{Generating Synthetic Training Data}
\label{sec:generation_pipeline}

Due to the lack of pretraining data, even SoTA LLMs' performance is unreliable when generating in an LRPL \cite{joel2024survey}. As we show in \S\ref{sec:datasets}, even GPT-4o scores only 58.2\% on a key Excel test set, showing this challenge. We must thus develop our method without requiring strong performance in the target domain by the teacher model. Fortunately, Excel (and other LRPLs) have rich, public documentation of language specifications and library functions. We leverage this (along with verification methods) to augment the teacher model and generate high-quality function demonstrations for the student model. We hypothesize that the student will learn to apply basic Excel functionality while transferring general problem-solving skills that it learned from pretraining in HRPLs.

\begin{figure*}[ht]
    \centering
    \includegraphics[width=1.0\linewidth]{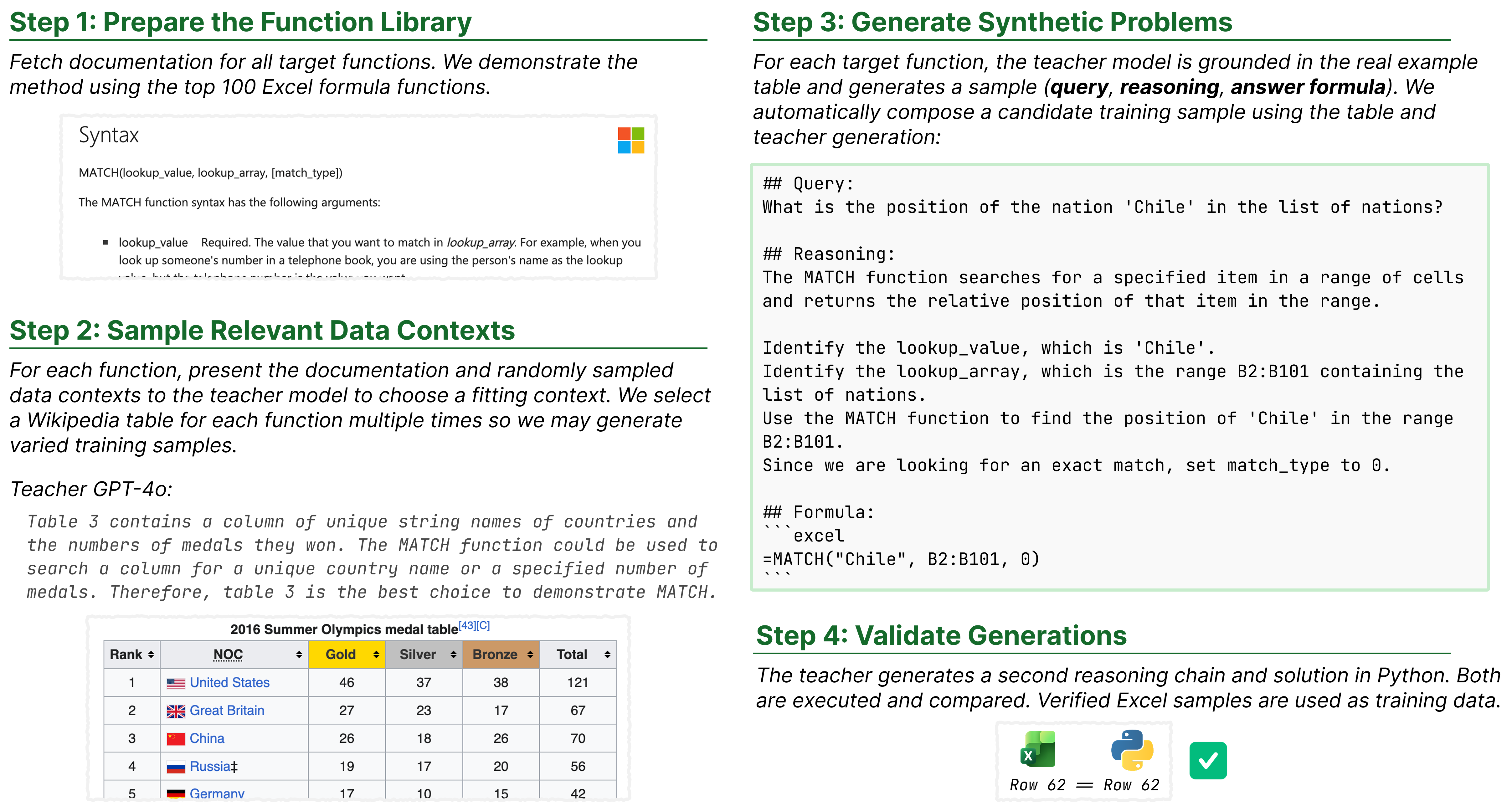}
    \caption{Our pipeline generates textbook-quality synthetic examples which demonstrate how to use target library functions. Examples are grounded in sampled, real data contexts, so the teacher model generates more creative problems and the student model learns from rich, varied demonstrations. We demonstrate producing a synthetic training example for the Excel MATCH function.}
    \label{fig:pipeline_demo}
\end{figure*}

Our pipeline only requires (a) LRPL function documentation, (b) unannotated, natural data contexts to serve as grounding, and (c) a teacher model which is generally strong at instruction-following (without requiring strong skills in the target LRPL domain). We demonstrate the method with Excel formulas, though it is suitable for any LRPL with these basic resources. It consists of 4 steps:

\subsection{Prepare the Function Library}
We first collect a library of target functions and their documentation. Excel Formula language contains 505 functions \citep{microsoftExcelFunctions}, yet many are unused due to niche functionality, or being superseded by newer implementations\footnote{Such as the newer XLOOKUP function which supersedes VLOOKUP and HLOOKUP.}. We next gather a small sample of public, unlabeled spreadsheets from the web, similarly to \citet{fisher2005euses}, and estimate a rough distribution of real function usage. We take the 100 most frequent functions and collect documentation pages from public Microsoft support.

For the function $f_i$ the documentation page $d_i$ contains a brief description of its purpose, its argument vector $\textbf{a}_i$ (including optional arguments), and their types. $d_i$ often also includes example usages of $f_i$ on a toy data table (a $\sim$5x5 table).

\subsection{Sample Relevant Data Contexts}
\cite{zhang2025llmsdesigngoodquestions} show that grounding synthetic generation in realistic data contexts aids model creativity and alignment to the target format. For simplicity and to demonstrate generality of the method, we seek easily accessible data. For Excel, WikiTableQuestions \citep{pasupat2015compositionalsemanticparsingsemistructured}, contains real data tables from Wikipedia; for other LRPLs different scenario data can be chosen. We randomly sample 10 Wikipedia tables per function from the training split. We query our teacher model GPT-4o to pick the best scenario table for each function, such that the function may be executable on the table. 
Manual inspection shows good quality: GPT-4o matches numeric functions to tables with numeric data, string functions to those with string data, etc.

\paragraph{Method details:} 
In the prompt to GPT-4o we insert the function $f_i$, its documentation page $d_i$, and 10 randomly selected tables.

In one sampling batch we acquire a random table for each of the 100 most-used functions in our target library and proceed to the next step with 100 (function, table) tuples. We repeat our entire pipeline until we have generated a suitable-sized curriculum; student models will thus train on a given function $f_i$ across many randomly sampled tables, to aid generalization of $f_i$ to new contexts.

\subsection{Generate Synthetic Problems}
We aim to generate synthetic demonstrations of function semantics by designing synthetic data samples which systematically show the effect of one function argument at a time. We query GPT-4o (denoted $Q(\cdot)$) to produce multiple synthetic problems from each input by demonstrating the target function $f_i$ with \textit{each} of its arguments in $\textbf{a}_i$ found in documentation $d_i$. All generations are grounded on the chosen table $t_i$: 
\begin{align*}
Q(f_i, d_i, t_i) \rightarrow S_i \\
S_i = \{(q, e, a)_j\}_{j \in 1..|\textbf{a}_i|}
\end{align*}

Even the teacher model may not be an expert in the LRPL, so $d_i$ provides the semantics of $f_i$, and the table $t_i$ provides grounding to generate natural questions. For one input, teacher GPT-4o is instructed to produce an output set of sample problems $z_j \in S_i$, where each $z_j = (q, e, a)$. The natural language question $q$ may be solved by executing $f_i$ on $t_i$ while making use of function argument $j$ of $f_i$. For each problem $z_j$ we also generate a step-by-step explanation $e$ of how to solve the problem, and an executable answer formula $a$. 

Rotating through each function argument and producing one demonstration for each allows us to demonstrate the usage of individual arguments and even optional ones. 

\paragraph{Method details:} We prompt the teacher model GPT-4o to generate textbook-quality tutorials in a QA format, demonstrating how to use a given function. A key aspect of the prompt is that we instruct the teacher to generate a set of examples, each one demonstrating how to use one argument slot of the target function. This is our method's most detailed and important prompt, so we share it in Figure~\ref{fig:prompt_generate_examples}. We elicit tutorial information in JSON, which allows us to extract structured information and compile the training tutorials automatically using a template. 

\begin{figure*}[!h]
\begin{tcolorbox}[colback=gray!10, colframe=black, width=\textwidth, title=Teacher Prompt: Generate Examples]
\begin{Verbatim}[fontsize=\scriptsize, breaklines=true]
## General Instruction:
You are a helpful assistant to a data scientist who is learning to use Excel. 

You are tasked with creating a tutorial of examples demonstrating the functionality of F, given F's reference documentation, as well as a random data table T taken from Wikipedia. 
The tutorial should contain at least one example demonstrating each of F's argument slots, in order to thoroughly describe how F works. 

## Task:
First, analyze the documentation of the function F to understand what each argument does. Write a brief explanation of what each argument is used for, including whether it is required or optional.
Format the explanation as markdown:

```markdown
Function: F
- arg1 <required>: explanation of arg1
- arg2 <required>: explanation of arg2
- arg3 <optional>: explanation of arg3
```

Second, write a series of examples demonstrating the use of F on the table T. Each example should contain:

1. The function F
2. The argument A being demonstrated
3. A natural language query Q which requires the use of F and A executed on the table T to compute a solution. Write the query in a natural and realistic way, as if an interested person were trying to analyze the data table to solve a problem. 
Make the query specific so there is only one correct answer. For example, to demonstrate a string manipulation function, the query Q should specify exactly how to format the output string so that a program can be written to do this.
4. A brief explanation of what F does in general (not related to the query Q or table T).
5. A step by step explanation of how to use F and A to solve the query Q given T. When explaining the steps, only use values mentioned in the query Q or references into the table T. Use the syntax section of the function F's documentation to explain how the arguments are used.
6. The answer to the query Q. After any reasoning, restate the answer on its own line at the end, e.g. "True", "False", "5", etc.
7. The final Excel formula using F and A to solve the query Q
8. Write the parameter name and required/optional for each of the final arguments given to F as a list, e.g. "param1 <required>", "param2 <optional>", etc.

Write examples which demonstrate the required arguments, then examples for each of the optional arguments. 
Format the examples as a JSON list according to the following structure:

```json
(...JSON description omitted for brevity...)
```

For the Excel formula, use the following format:
"=FUNCTION(ARGUMENTS)"

## Function:
MATCH

## Documentation:
(...MATCH documentation omitted for brevity...)

## Random Table:
In Excel tables, the first row is usually reserved for column headers. The first column is usually reserved for row headers. For example, the data starts in A2.
Larger tables may be excerpted here. If so, the first and last rows of the table will be shown, with an ellipsis (...) in between representing the hidden middle rows.
Remember that NaN values in Excel may be written in the table as "nan".

(...table omitted for brevity...)

## Tutorials:
    \end{Verbatim}
    \end{tcolorbox}
    \caption{We instruct teacher GPT-4o to generate multiple tutorials for a target function MATCH, demonstrating each function argument. We add minimal details about Excel solely to describe the data format such as the excerpting process for large tables. The method can be tailored easily to a target LRPL.}
    \label{fig:prompt_generate_examples}
\end{figure*}

\subsection{Validate Generations}

Each generated problem demonstrates in textbook-quality the basic function semantics in the target domain, with natural language explanation. However, since samples are synthetically generated using an AI model, they are unverified for correctness. We improve data quality by employing post-generation validation. First, we filter out all samples with answer formula $a$ which fail to execute. Second, we feed generated samples back into the teacher model and generate parallel solutions in Python to the same questions. We do not know the correct answer to synthetic questions without human labeling, but we assume Python (as an HRPL) will have the most reliable generations \cite{joel2024survey}. Indeed, we find roughly a 50\% match rate between Excel and Python executed values, and retain the verified samples.

We show that this validation is useful. We finetune a Qwen2.5-Coder 14B model on equal amounts of validated and un-validated data (6,440 samples in each) per the procedure in \S\ref{ssec:finetuning}. We observe that when trained on unvalidated data, the model improves from 46.95\% to 52.11\% on the downstream WikiTQ test set used in experiments (\S\ref{sec:datasets}), compared to 54.93\% when finetuned on an equal amount of validated data. This demonstrates that (a) unvalidated data has great potential even with just simple filtering out of non-executable samples, and (b) if available, parallel generation of solutions in an HRPL like Python can provide useful supervision for higher-quality assurance, resulting in downstream task improvement.


\paragraph{Method details:} The teacher model generates sample problems with Excel reasoning and solutions. A problem generated by the teacher model forms a tuple: (table, query, Excel formula, Excel execution) (we execute the formula ourselves for reliability). We format tuples into training data after filtering them for quality. When filtering we generate a parallel solution to each problem in Python, a high resource programming language in which we expect model solutions to be of high quality. We give nearly the same prompt as in model evaluations, but cast each sample as a Python problem using a Pandas DataFrame instead of an Excel spreadsheet. We collect the generated Python code and execute it so that we may compare the Python and Excel executed values.

Because Pandas DataFrames have subtle differences to spreadsheets (e.g. they are 0-indexed whereas Excel is 1-indexed), we develop some basic rules for determining equality between executed values. We are able to accept around 50\% of synthetic generations after Python-validation.

\begin{table*}[t]
    \footnotesize
    \centering
    \begin{tabular}{l|rrrrrr}
    \toprule
        & Base Model & $RAG_{All}$ & $RAG_{Oracle}$ & $FT_{Doc}$ & $FT_{Doc-QA}$ & $FT_{Syn-QA}$ \\ \midrule
        GPT-4o & 79.19 & 78.25 & \textbf{84.19} & - & - & - \\
        \midrule
        Qwen2.5-coder 3B & 15.34 & 13.77 & 18.94 & 13.62 & 15.34 & \textbf{28.64} \\
        Qwen2.5-coder 14B & 46.95 & 44.76 & 53.52 & 40.22 & 46.95 & \textbf{54.93} \\[0.2cm]
        Qwen2.5 3B & 14.87 & 12.21 & 17.84 & 13.62 & 14.55 & \textbf{20.81} \\
        Qwen2.5 14B & 50.70 & 47.57 & 49.45 & 46.64 & 49.92 & \textbf{51.02} \\
        \midrule
        CodeLlama-Instruct 7B & 0.47 & 0.31 & 0.47 & 0.78 & \textbf{9.23} & 4.85 \\
        CodeLlama-Instruct 13B & 10.80 & 13.15 & 16.12 & 12.21 & 12.83 & \textbf{21.28} \\[0.2cm]
        Llama2 7B & 0.63 & 0.16 & 1.10 & 0.63 & 3.91 & \textbf{8.61} \\
        Llama2 13B & 1.72 & 0.31 & 1.56 & 0.47 & 4.54 & \textbf{15.49} \\
        \bottomrule
    \end{tabular}
    \caption{Evaluation results on our subset of WikiTableQuestions. Execution Match (EM) measures the percentage of programs which execute to the correct answer. We test (a) base models; (b) RAG settings: all function signatures ($RAG_{All}$) and oracle-retrieved signatures ($RAG_{Oracle}$); and (c) finetuned models: using function documentation ($FT_{Doc}$), and QA-formatted documentation ($FT_{Doc-QA}$), and synthetic problems ($FT_{Syn-QA}$). Finetuning on synthetic problems performs better than oracle-retrieved RAG or finetuning on purely documentation-based data.}
    \label{tab:results_wtq}
\end{table*}

\section{Experiment: Excel Generation}
\label{sec:experiments}

We augment models in several ways and compare performance in Excel formula generation.

\subsection{Example Domain and Datasets}
\label{sec:datasets}

LRPLs lack publicly available data. Excel, while popular, is mainly used in enterprise settings where tabular data is sensitive and private, making it rare to find real-world examples that pair human-written Excel formulas with tables, and natural language queries. Thus, it is an LRPL.
Accordingly, we recast existing table-based QA datasets into the Excel domain by embedding tables into Excel spreadsheets and prompting models to answer using Excel formulas:
\begin{itemize}
    \item \textbf{WikiTQ}: WikiTableQuestions \cite{pasupat2015compositionalsemanticparsingsemistructured} contains tables from Wikipedia annotated by humans with questions and answers in natural language.
    \item \textbf{TAT-QA}: A financial table-QA dataset in a similar format as WikiTQ, and in a domain relevant to Excel generation \cite{zhu-etal-2021-tat}.
\end{itemize}

To ensure these problems are solvable using Excel, we use OpenAI GPT-4o and o1 \cite{openai2024openaio1card} to attempt a solution to each problem using Excel, and we collect problems which are solvable in this domain.
On WikiTQ cast into Excel, GPT-4o scores 58.2\% and o1 scores 67.8\%. We cannot know if o1's score being far below 100\% is due to either model incapability in the Excel domain or problem incompatibility when cast into Excel. However, GPT-4o's score below o1's \textit{must} be due to model incapability, demonstrating that it is not strong in our target LRPL domain off-the-shelf.

Therefore, we take the subset of samples passed by the stronger o1 as our test dataset, since they are proven to be solvable in the Excel domain, and we use the o1 formulas as the \textbf{oracle} Excel solution for each sample. This results in 639 problems for our \textbf{WikiTQ} test and 459 for \textbf{TAT-QA} test. In experiments we evaluate model performance using program execution match (EM) to executed values and do not grade the generated formulas; we only use the oracle formulas for a baseline comparison.

\subsection{Student Models}\label{ssec:model_names} 


We select student model families that are (1) open-weights, (2) available in multiple sizes, and importantly, (3) have corresponding code-finetuned versions of generally trained models.

\begin{itemize}
    \item \textbf{Qwen 2.5} A recent family with notable performance across tasks, including analytical reasoning and coding \cite{qwen2025qwen25technicalreport}. We use Qwen2.5 3B and 14B; and Qwen2.5-Coder 3B and 14B.
    \item \textbf{Llama 2} An older, but popular family with corresponding code-finetuned models \cite{touvron2023llama2openfoundation}. We use Llama2 7B and 13B; and CodeLlama 7B and 13B.
\end{itemize}

\subsection{Student Model Augmentation}

We compare several student model augmentations which broadly fall into two categories: off-the-shelf RAG methods and finetuning methods.

\subsection{RAG}
Retrieval-Augmented Generation is a cheap and effective technique to supply novel information to a model by directly inserting textual knowledge from external sources into the model prompt \citep{lewis2020rag}.
We simulate baselines for both na\"ive and optimal RAG scenarios in which LRPL function library data is given to the model. 
\begin{itemize}
    \item In the \textbf{RAG$_{All}$} setting, we provide student models with all of the 100 most frequently used Excel function\footnote{The same 100 functions as in finetuning experiments.} signatures and their descriptions before the question. However, this may naturally create a difficult needle-in-the-haystack scenario where information gets lost.
    \item \textbf{RAG$_{Oracle}$} is an expected upper bound in which we augment with \textit{only} the functions included in the o1 oracle Excel solution for each question. This simulates an optimal retriever and minimizes the context a student model must consider.
\end{itemize}

\subsection{Fine-tuning}
\label{ssec:finetuning}
We compare several finetuning scenarios demonstrating common approaches. Our method is:

\begin{itemize}
    \item \textbf{FT$_{Syn-QA}$}: finetuning student models on our synthetic demonstration data. 
\end{itemize}

For \textbf{FT$_{Syn-QA}$} we generate 6,440 validated samples using our pipeline (\S\ref{sec:generation_pipeline}), which demonstrate the top 100 Excel functions and their argument semantics. We then finetune student models on this data. Models train to convergence on a heldout set of 100 questions from WikiTQ dev, typically 1-3 epochs. 
We finetune all models using the hyperparameters shown in Table~\ref{tab:hyperparameters}. We chose these by doing hyperparameter search over the given values using Qwen2.5-coder 3B and choosing the parameters which yielded the highest score on the heldout validation dataset. We then used these settings to train all our models, and applied these finetuned models to the test sets.

\begin{table}[tb]
    \centering
    \footnotesize
    \begin{tabular}{lll}
        \toprule
        Hyperparameter & Value & Search Range \\ \midrule
        Batch size & 4 & \{2, 4, 8, 16, 32\} \\
        Learning Rate & 5e-5 & \{1e-5, 5e-5, 1e-4, 5e-4\} \\
        LoRA r & 64 & \{32, 64, 128\} \\
        LoRA $\alpha$ & 1 & \\
        Max epochs & 6 & \\
        Patience & 3 & \\
        LR scheduler & cosine & \{linear, cosine\} \\
        Warmup ratio & 0.5 & \\ \bottomrule
    \end{tabular}
    \caption{Hyperparameter choices for finetuning.}
    \label{tab:hyperparameters}
\end{table}

We also test two finetuning baselines:

\begin{itemize}
    \item \textbf{FT$_{Doc}$} tests a ``continued training'' scenario: models finetune on the raw documentation for the 100 most used Excel functions, which contain simple examples paired with toy tables. However, the documentation format used (including toy examples) does not mirror the format of the downstream QA task, which may be suboptimal for the student model.

    \item \textbf{FT$_{Doc-QA}$} addresses format by using GPT-4o to reformat extracted examples from the documentation into the same QA format of the downstream task. This involves purely restructuring natural language content only, and no generation of novel, synthetic content.
\end{itemize}


We also use the \textbf{FT$_{Doc-QA}$} baseline as a control for understanding our synthetic training data: because the format is the same,
improvement over this baseline must be attributed to the \textit{content} which we generate using our method. 



\section{Results and Discussion}
\subsection{WikiTQ} Table~\ref{tab:results_wtq} shows full results on WikiTQ.

In \textit{$RAG_{All}$} all models except for CodeLlama 13B suffer when given all function signatures, likely because the long context obscures the most relevant function. However, optimal retrievals in \textit{$RAG_{Oracle}$} improve most models, showing the upper-bound of a RAG approach. However, improvement is inconsistent and not large over \textit{Base Models}.

In the finetuned \textit{$FT_{Doc}$} setting,``continued training'' on raw function docs often harms performance. However, restructuring this into QA format (\textit{$FT_{Doc-QA}$}) can sometimes be useful, such as in the weaker LLama2 models. Yet compared to \textit{Base Models} the improvement is also marginal.

In \textit{$FT_{Syn-QA}$} we achieve a significant performance boost over all \textit{Base Models} by finetuning on our synthetic data. The finetuned models even outperform corresponding \textit{$RAG_{Oracle}$} settings, where the functions expected to be used are provided as hints by oracle o1. See Figure~\ref{fig:hero} for a showcase. 

Regarding code pretraining, it is not clear that base models pretrained in HRPLs exhibit any consistent improvement in generating Excel, an LRPL.
However, most code-pretrained models enjoy a greater boost from synthetic data finetuning compared to non-specialized counterparts. This shows clearly that previously learned coding skills in HRPLs can transfer to LRPLs.
 

\subsection{TAT-QA}
We show results of evaluating base models and finetuned models on TAT-QA in Table~\ref{tab:results_tat_qa}. 

\begin{table}[tb]
    \centering
    \footnotesize
    \begin{tabular}{l|rr}
        \toprule
        ~ & Base Model & $FT_{Syn-QA}$ \\ \midrule
        GPT-4o & 77.78 & - \\ \midrule
        Qwen2.5-coder 3B & 5.88 & \textbf{8.06} \\
        Qwen2.5-coder 14B & 14.37 & \textbf{14.60} \\[0.2cm]
        Qwen2.5 3B & 6.75 & \textbf{11.11} \\
        Qwen2.5 14B & \textbf{15.47} & 15.03 \\ \midrule
        CodeLlama-Instruct 7B & 0.44 & \textbf{2.18} \\
        CodeLlama-Instruct 13B & 3.49 & \textbf{7.41} \\[0.2cm]
        Llama2 7B & 0.00 & \textbf{3.92} \\
        Llama2 13B & 1.09 & \textbf{6.54} \\ \bottomrule
    \end{tabular}
    \caption{Evaluation results on our subset of TAT-QA. Execution Match (EM) results are shown.}
    \label{tab:results_tat_qa}
\end{table}

We observe similar trends on TAT-QA as we do on WikiTQ. Large models perform better than small models, and our finetuning data is able to improve 7 of 8 models significantly. We notice that in general, the effect size of finetuning is smaller when evaluated on TAT-QA, and we look to the lower base model performances to explain this. While TAT-QA is a similarly-formatted dataset in which data tables are paired with natural language queries, it is in a very specialized domain. Financial data is likely to be more scarce in general LLM pretraining data, so we expect a priori that performance may be lower on this dataset, and the lower base model performances compared to WikiTQ confirm this.

We also note that it is unclear if code-specialized models outperform non-code-specialized models on this task. We believe this is for the same reason as above: code-specialization does not appear to improve models' financial reasoning ability, and our further code-finetuning in the general domain of Wikipedia tables may not show particular benefits to coding models on this dataset, either.

\subsection{Error Analysis of Model Generations}
\label{ssec:analysis}

We randomly sample 100 problems from the WikiTQ test set and conduct a qualitative analysis of the model generations produced by Qwen2.5-Coder 3B \textit{Base Model} and \textit{$FT_{Syn-QA}$}.\footnote{This code-pretrained model was chosen for its large performance boost in testing.}
Table~\ref{tab:analysis_categories} demonstrates that fine-tuning on synthetic data primarily reduces errors in table reading (-22 questions) and formula syntax (-8), while preserving the model's ability in planning and function selection. The small increase in planning errors is attributed to the model's tendency to produce single-function formulas. Since the synthetic data focuses on the use of individual functions, we observe that 64.3\% of the solutions generated by \textit{$FT_{Syn-QA}$} involve a single function, compared to 44.4\% in the base model. 
In our error analysis we use the categorization guidelines in Table~\ref{tab:analysis_categories_guidelines} to analyze the generations from Qwen models on our WikiTQ test set.

\begin{table}[tb]
    \centering
    \footnotesize
    \begin{tabular}{lrr}
        \toprule
        Result Category & Base Model & $FT_{Syn-QA}$ \\ \midrule
        Correct & 13 & 34 \\ \midrule
        Error: Plan Logic & 25 & 33 \\
        Error: Function Choice & 8 & 9 \\
        Error: Table Indexing & 35 & 13 \\
        Error: Formula Syntax & 9 & 1 \\
        Error: Other & 10 & 10 \\
        \bottomrule
    \end{tabular}
    \caption{Analysis of errors by Qwen2.5-Coder 3B base and finetuned on 100 WikiTQ test samples.}
    \label{tab:analysis_categories}
\end{table}

\begin{table}[tb]
    \centering
    \footnotesize
    \begin{tabular}{p{0.4\linewidth}p{0.52\linewidth}}
        \toprule
        Result Category & Guideline \\ \midrule
        Correct & {\scriptsize The model generation executes to the correct answer} \\ \midrule
        Error: Plan Logic & {\scriptsize If the CoT is wrong} \\
        Error: Function Choice & {\scriptsize If the CoT is correct, but it decides to use the wrong function, or is missing some functions} \\
        Error: Formula Syntax & {\scriptsize If the function is correct, but the way of using the function is wrong} \\
        Error: Table Indexing & {\scriptsize If the function usage is correct, and changing the cell/row/column numbers would get the correct answer} \\
        Error: Other & {\scriptsize Any other error, such as correct plan but incorrect formula semantics} \\
        \bottomrule
    \end{tabular}
    \caption{Annotation guide for error classification.}
    \label{tab:analysis_categories_guidelines}
\end{table}

\begin{table}[tb]
    \centering
    \footnotesize
    \begin{tabular}{l|rr}
        \toprule
        ~ & Base Model & $FT_{Syn-QA}$ \\ \midrule
        Qwen2.5-coder 3B & 44.44 & 64.32 \\
        Qwen2.5-coder 14B & 46.17 & 59.00 \\[0.2cm]
        Qwen2.5 3B & 40.22 & 73.08 \\
        Qwen2.5 14B & 40.22 & 40.85 \\ \bottomrule
    \end{tabular}
    \caption{Percentage of generated formulas consisting of a single function.}
    \label{tab:single_func}
\end{table}

\begin{table}[tb]
    \centering
    \footnotesize
    \begin{tabular}{lrr}
        \toprule
        ~ & Improved & Regressed \\
        \midrule
        Qwen2.5-coder 3B & 29.77 & 10.87 \\
        Qwen2.5-coder 14B & 27.27 & 7.14 \\[0.2cm]
        Qwen2.5 3B & 23.66 & 3.64 \\
        Qwen2.5 14B & 17.33 & 21.92 \\ \bottomrule
    \end{tabular}
    \caption{Improvements and regressions when the student model predicts the same, single-function formula before and after finetuning. Models master individual functions and often improve in generating single-function calls, usually with little regression.}
    \label{tab:improvement_func}
\end{table}

Next, we turn to analyzing model performance before and after our function finetuning. We first show the percentage of model-generated formulas which consist of a single function call in Table~\ref{tab:single_func}. We observe that all models increased the number of single function predictions after finetuning, however this effect is more noticeable in smaller models.

Shown in Table~\ref{tab:improvement_func}, of each model's improvements (samples which were incorrect before tuning, and correct after), many generations produce the same function before and after. This shows learned mastery of these functions, where finetuned models produce better formulas using the same function. We also observe minimal regressions when predicting the same function (correct usage before tuning, but incorrect after).

\section{Conclusion}
We show that generating textbook-quality examples for a low resource programming language can rapidly improve LLM generation through finetuning, with benefits over RAG approaches and other, less rich finetuning paradigms. Yet our method is simple and automatic, requiring only function documentation, a teacher model generally strong in instruction-following (which need not be an expert in the target domain), and minimal human description of the domain. We demonstrate the method for Excel Formula language which is popular in practice, but an LRPL owing to its sparsity of training data. Our method improves student models by leveraging their problem solving abilities previously learned in pretraining on high-resource programming languages to adapt to the new domain.

We posit that further work on individual languages may benefit similarly, and we encourage future work in synthetic data generation for LLM pretraining, especially for ``long-tail'' phenomena like LRPLs which suffer due to a lack of natural training data.

\section*{Limitations}

We demonstrated our methods on a single LRPL, though our findings should apply to any low-resource programming language due to the generality of our method. We also focused on generating demonstrations for just the function library in our example LRPL, a more realistic target for generating textbook-quality data than the infinite space of possible programs, especially when limited resources exist in the target domain. While we cannot compare to the challenges or benefits of generating complex data, as we noted in error analysis this limitation clearly biases models towards simpler solutions, so this presents a problem which future research may alleviate. Further, due to computational costs we only investigated methods for pruning bad synthetic generations, and not any methods for fixing those bad generations. Research in this direction could also prove valuable.

\nocite{*}
\section{Bibliographical References}\label{sec:reference}

\bibliographystyle{lrec2026-natbib}
\bibliography{custom}


\bibliographystylelanguageresource{lrec2026-natbib}

\end{document}